\documentclass[acmtog]{acmart}

\usepackage{graphicx}
\usepackage{subfig}
\usepackage{soul}
\usepackage{algorithmicx}
\usepackage{multirow}

\usepackage{mathtools}
\usepackage{bm} 
\usepackage{algorithm}
\usepackage{algpseudocode}
\usepackage{enumitem}

\usepackage{pifont}
\usepackage{xspace}
\newcommand{\ssname}{ShapeShift\xspace}

\usepackage{xcolor}
\newif\ifshowauthorcomment

\renewcommand\footnotetextcopyrightpermission[1]{}
\begin{document}

\begin{teaserfigure}
\centering%
   \includegraphics[width=\linewidth]{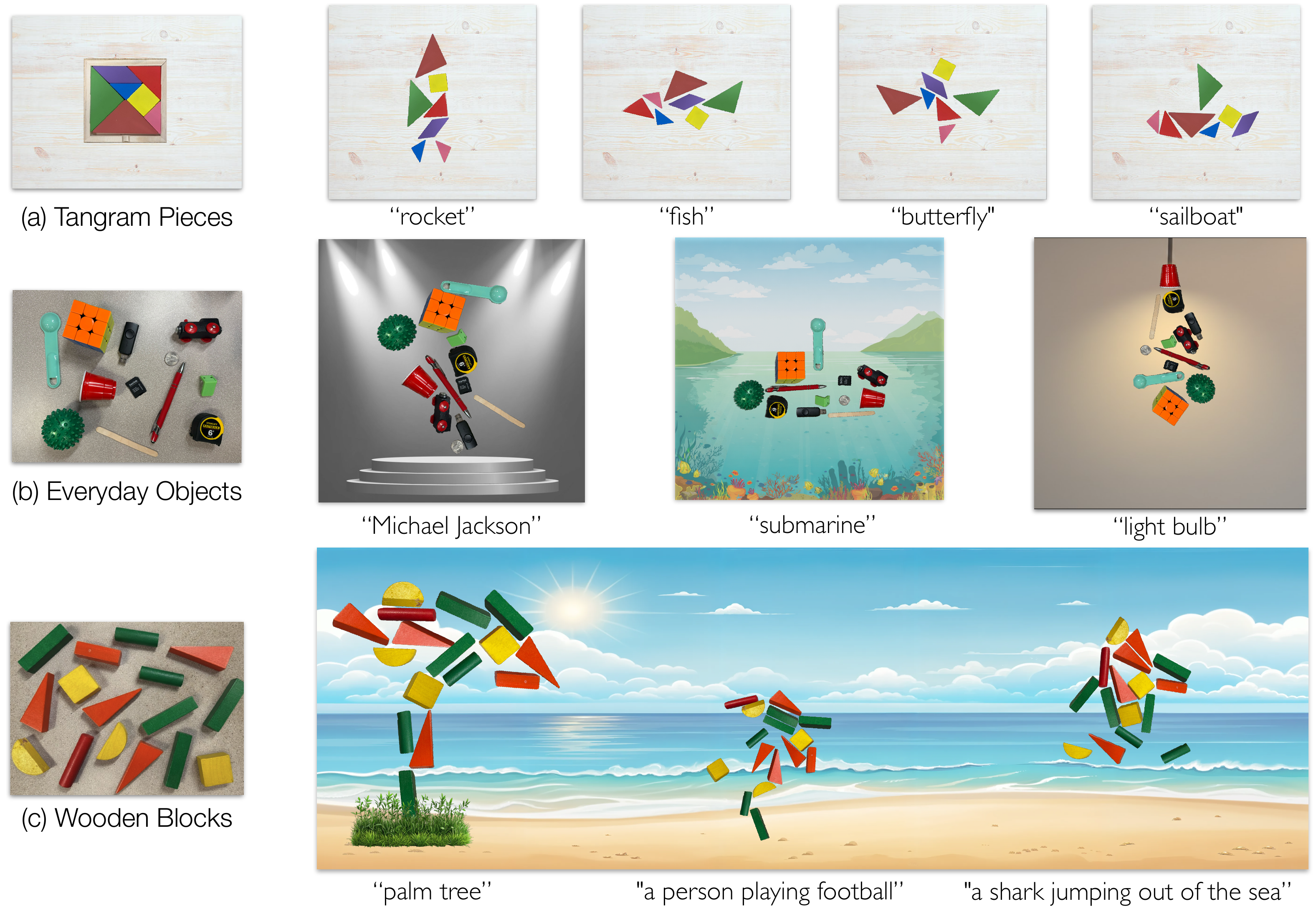} \vspace{-17pt}
        \captionof{figure}{ShapeShift transforms rigid objects into semantically meaningful arrangements from natural language. Given input shapes—(a) tangram pieces, (b) everyday objects, or (c) wooden blocks—and a text prompt such as ``rocket'' or ``a person playing football,'' our method discovers overlap-free configurations where the shapes collectively depict the described concept. Unlike image generation, the output uses only the provided shapes with physically valid, non-overlapping placements. Results demonstrate generalization across geometric primitives and irregular real-world objects.}%
        \label{fig:teaser}%
\end{teaserfigure}

\author{Vihaan Misra}
\affiliation{%
  \institution{Carnegie Mellon University}
  \country{USA}
}

\author{Peter Schaldenbrand}
\affiliation{%
  \institution{Carnegie Mellon University}
  \country{USA}
}

\author{Jean Oh}
\affiliation{%
  \institution{Carnegie Mellon University}
  \country{USA}
}

\email{{vihaanm, pschalde, hyaejino}@andrew.cmu.edu}

\renewcommand{\shortauthors}{Misra et al.}

\title{ShapeShift: Text-to-Mosaic Synthesis via Semantic Phase-Field Guidance}



\begin{abstract}
We present \ssname, a method for arranging rigid objects into configurations that visually convey semantic concepts specified by natural language. While pretrained diffusion models provide powerful semantic guidance, such as Score Distillation Sampling, enforcing physical validity poses a fundamental challenge. Naive overlap resolution disrupts semantic structure, e.g., separating overlapping shapes along geometrically optimal directions--minimum translation vectors--often destroys the very arrangements that make concepts recognizable.
Our intuition is that diffusion model features encode not just what a concept looks like, but its geometric, directional structure—how it is oriented and shaped—which we leverage to make overlap resolution semantically aware.  We introduce a deformable boundary represented as a phase field that expands anisotropically, guided by intermediate features from the diffusion model, creating space along semantically coherent directions. Experiments demonstrate that ShapeShift, by coupling semantic guidance and feasibility constraint resolution, produces arrangements achieving both semantic clarity and overlap-free validity, significantly outperforming baselines that treat these objectives independently.
\end{abstract}

\maketitle
    
\section{Introduction} \label{sec:intro}

The human ability to create meaning from simple arrangements is remarkable, as demonstrated by tangram puzzles and toy blocks, where a small set of rigid pieces can be composed into countless recognizable configurations. Producing such arrangements requires simultaneously capturing semantic intent and respecting physical constraints---a capability that remains challenging for artificial systems despite recent advances in generative models.

\begin{figure}
    \centering
    \includegraphics[width=\linewidth]{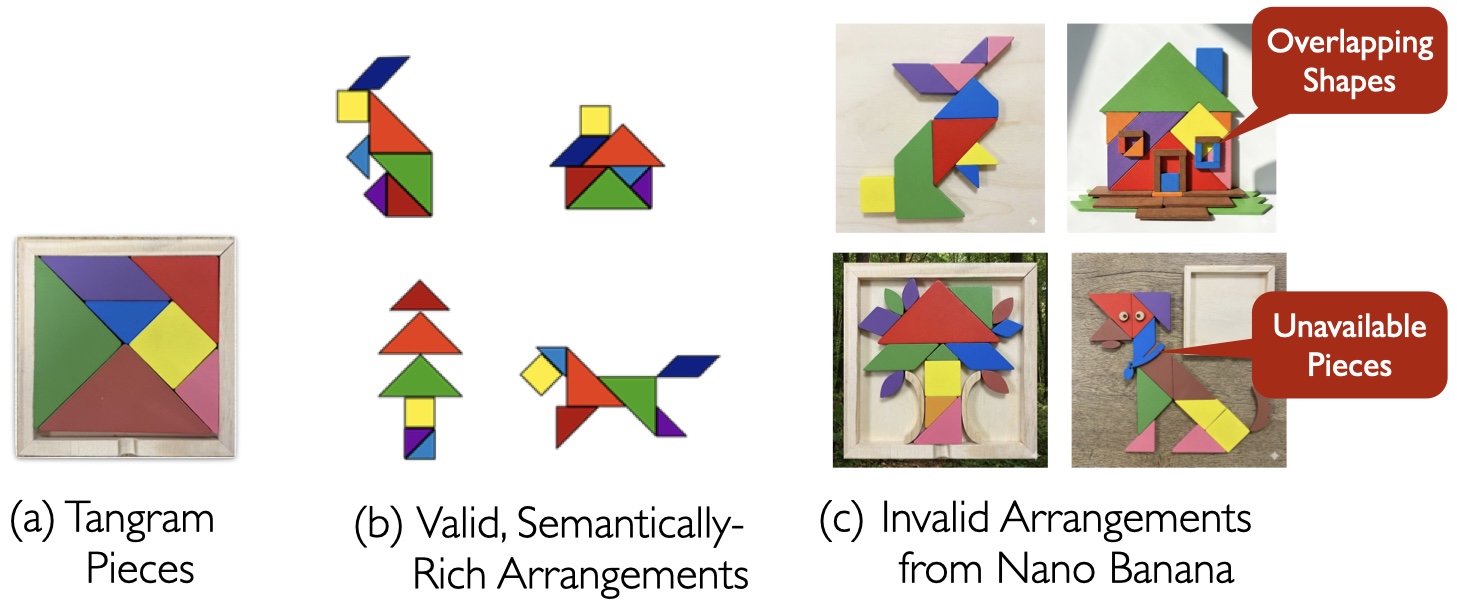}
    \caption{\textbf{The Gap Between Generation and Arrangement.} Humans create semantically rich tangram arrangements using fixed geometric pieces. 
    Nano Banana Pro~\cite{geminiteam2025geminifamilyhighlycapable} generates visually plausible tangrams but violates physical constraints—using unavailable pieces or producing impossible overlaps. This illustrates the challenge we address: grounding semantic generation with explicit geometric constraints.}
    \label{fig:motivation2}
\end{figure}

Modern image generation models~\cite{esser2024stable-diffusion3,Rombach22,Saharia22} excel at generating diverse imagery from natural language, but operate in continuous pixel or latent spaces without understanding physical constraints. As Fig.~\ref{fig:motivation2} illustrates, generates tangram arrangements that are visually appealing but that violate the rules of the game, i.e., use nonexistent pieces, modify shapes, or introduce overlaps. 

In this context, we introduce the \emph{text-to-mosaic} task, a generalized variant of the tangram puzzle: given a set of rigid objects and a text description, rearrange the objects to visually convey the specified concept under three rules: (1)~object geometry (shape and size) must not be modified; (2)~all pieces must be used; (3) the identities of the pieces must be preserved; and (4)~objects must not overlap.
We formulate text-to-mosaic as optimization over object poses: find translations and orientations that maximize semantic alignment to the prompt while satisfying geometric feasibility.

In this paper, we formulate text-to-mosaic arrangement as optimization: find positions and orientations for each object that jointly maximize semantic alignment while satisfying geometric constraints. 

A natural source of semantic guidance is Score Distillation Sampling (SDS)~\cite{poole2022dreamfusiontextto3dusing2d}, which provides gradients from a pretrained diffusion model. However, naively combining SDS with plain overlap resolution yields poor results (Fig.~\ref{fig:baselines}).  Generally, separating overlapping shapes can inadvertently  degrade the semantic quality of the arrangement, and algorithms that perform geometrically optimal separations can destroy the very structure that makes a concept recognizable without semantic knowledge.
For instance, consider arranging triangular shapes into a sword as seen in Fig.~\ref{fig:baselines}. SDS discovers that blade shapes should align along an elongated axis, but these shapes overlap. Standard overlap resolution computes minimum translation vectors (MTVs), pushing shapes apart along the shortest escape direction. For blade shapes, this direction is typically perpendicular to the blade axis, causing shapes to spread \emph{across} the blade rather than \emph{along} it. Repeated application transforms an elegant sword into an amorphous blob. The problem is not that overlap resolution fails geometrically—it succeeds—but that it fails \emph{semantically} as illustrated in~Fig.\ref{fig:baselines}.

As opposed to separating the two objectives, we suggest that \textbf{overlap resolution and semantic preservation \emph{should} be coupled}.
%
We introduce a \emph{diffuse-interface (phase-field) membrane} \(u:\Omega\!\to\![0,1]\)---a soft container whose \(u{=}0.5\) level set defines the feasible region boundary---and we evolve this membrane using semantic structure extracted from a diffusion model's intermediate UNet features. Prior work shows that such intermediate diffusion features can encode useful spatial structure (e.g., semantic correspondence)~\cite{Tang2023DIFT}. Leveraging this signal, we expand the membrane \emph{anisotropically}: in regions where features indicate coherent structure, the membrane preferentially expands along that structure rather than across it. Overlap resolution then operates within these semantically informed bounds, improving feasibility while preserving and improving the discovered semantic layout.

The membrane's evolution is governed by an anisotropic energy functional where the diffusion tensor is derived from the structure of UNet features. In regions where features indicate elongated semantic structure (e.g., a blade), the membrane preferentially expands along that structure rather than across it. Overlap resolution then operates within these semantically-informed bounds, achieving physical validity while preserving semantic meaning.

Our method, \emph{ShapeShift}, operates in two phases. \textbf{Phase~1} performs asset-conditioned semantic discovery via multi-scale SDS, allowing overlaps so shapes can organize into a concept-consistent configuration. \textbf{Phase~2} enforces feasibility while preserving this structure: we iteratively project poses toward non-overlap inside a semantically-guided membrane that adapts to create space primarily along concept-consistent directions.

\vspace{-0.25em}
\paragraph{Contributions.}
\vspace{-0.35em}
\begin{enumerate}
    \setlength\itemsep{0.15em}
    \setlength\parskip{0pt}
    \setlength\topsep{0.25em}
    \item We identify the fundamental tension between overlap resolution and semantic preservation in constrained arrangement synthesis, showing that naive approaches fail because geometric optimality conflicts with semantic structure.
    
    \item We introduce the semantic phase-field membrane, a deformable boundary whose anisotropic evolution is guided by diffusion model features, enabling overlap resolution that preserves semantic meaning.
    
    \item We demonstrate through extensive evaluation, including human studies, that ShapeShift significantly outperforms baselines in both semantic accuracy and physical validity, producing arrangements that are both meaningful and overlap-free.
\end{enumerate}
\vspace{-0.35em}

\section{Related Work}
\label{sec:related}

\paragraph{Differentiable rendering and text-guided optimization.}
Differentiable rasterizers such as DiffVG~\cite{Li20} enable gradient-based optimization over vector-graphics primitives, and have been used for text-driven drawing and SVG synthesis~\cite{frans2021clipdrawexploringtexttodrawingsynthesis,schaldenbrand2022styleclipdrawcouplingcontentstyle,iluz2023wordasimagesemantictypography,jain2022vectorfusiontexttosvgabstractingpixelbased}. In parallel, Score Distillation Sampling (SDS)~\cite{poole2022dreamfusiontextto3dusing2d} leverages pretrained diffusion priors~\cite{ho2020denoisingdiffusionprobabilisticmodels,song2022denoisingdiffusionimplicitmodels,Rombach22} to optimize external parameters (e.g., 3D geometry, vector graphics, or layouts) toward a text prompt. Our work builds on these optimization-based paradigms, but differs in \emph{what is optimized} and \emph{what must be enforced}: we optimize only rigid poses of fixed input shapes and explicitly enforce non-overlap and containment, which are not addressed in most text-driven vector-graphics or SDS-based synthesis settings.

\paragraph{Arrangement Prediction.}
Recent work leverages language and vision-language models for spatial arrangement. Dream2Real~\cite{kapelyukh2024dream2realzeroshot3dobject} and StructDiffusion~\cite{liu2022structdiffusion} predict object placements from high-level goals, but typically handle simple configurations only (``set the table,'' ``put objects in a line''). Dall-E-Bot~\cite{kapelyukh2023dall-e-bot} uses image generation to guide rearrangement but requires distinct, nameable objects and fails with ambiguous shapes like wooden blocks. Diffusion-based models have also been explored for layout generation in discrete design spaces, enabling controllable placement of elements under a generative prior~\cite{Inoue_2023_CVPR_LayoutDM}. Closer to our optimization setting, Lay-A-Scene leverages text-to-image priors at test time to infer object arrangements, illustrating the strength of diffusion priors for spatial composition even when the objects are fixed~\cite{Rahamim2024LayAScene}. However, these systems are not designed for the \emph{text-to-mosaic} setting where (i) objects are \emph{fixed, unlabeled shapes} without per-object semantic descriptions, (ii) \emph{all pieces must be used}, and (iii) \emph{strict non-overlap} is required. Related systems for assembling or rearranging primitives often rely on restricted shape families or fixed part libraries: Blox-Net~\cite{goldberg2024bloxNet} and BrickGPT~\cite{pun2025generatingphysicallystablebuildable} rearrange 3D primitives but are limited to simple primitive types or a predefined brick set, respectively. In contrast, ShapeShift arranges arbitrary 2D shapes into open-vocabulary concepts while explicitly enforcing geometric feasibility and preserving semantic coherence.

\paragraph{Overlap detection, packing, and differentiable non-penetration.}
Classical overlap detection relies on geometric tests such as the Separating Axis Theorem (SAT)~\cite{sat,Lindemann2009TheGDgjk} for convex polygons, with resolution typically via minimum translation vectors (MTVs)~\cite{Ericson05}. Recent learning-based layout systems incorporate collision avoidance (e.g., LayoutVLM~\cite{sun2024layoutvlm}), but generally resolve intersections in geometric space without semantic context, which can produce feasible yet semantically degraded configurations. In parallel, other work has developed \emph{differentiable} non-overlap / non-penetration constraints for optimization, including Minkowski-style penalty formulations~\cite{Minarcik2024MinkowskiPenalties} and analytic differentiable collision conditions~\cite{Jaitly2025AnalyticCollision}. We leverage differentiable overlap gradients for pose projection, but our key contribution is orthogonal: we make overlap resolution \emph{semantically informed} by using diffusion-model features to determine \emph{where} and \emph{along which directions} feasibility restoration should occur.

\section{Method}
\label{sec:method}

\begin{figure*}[t]
    \centering
    \includegraphics[width=0.8\textwidth]{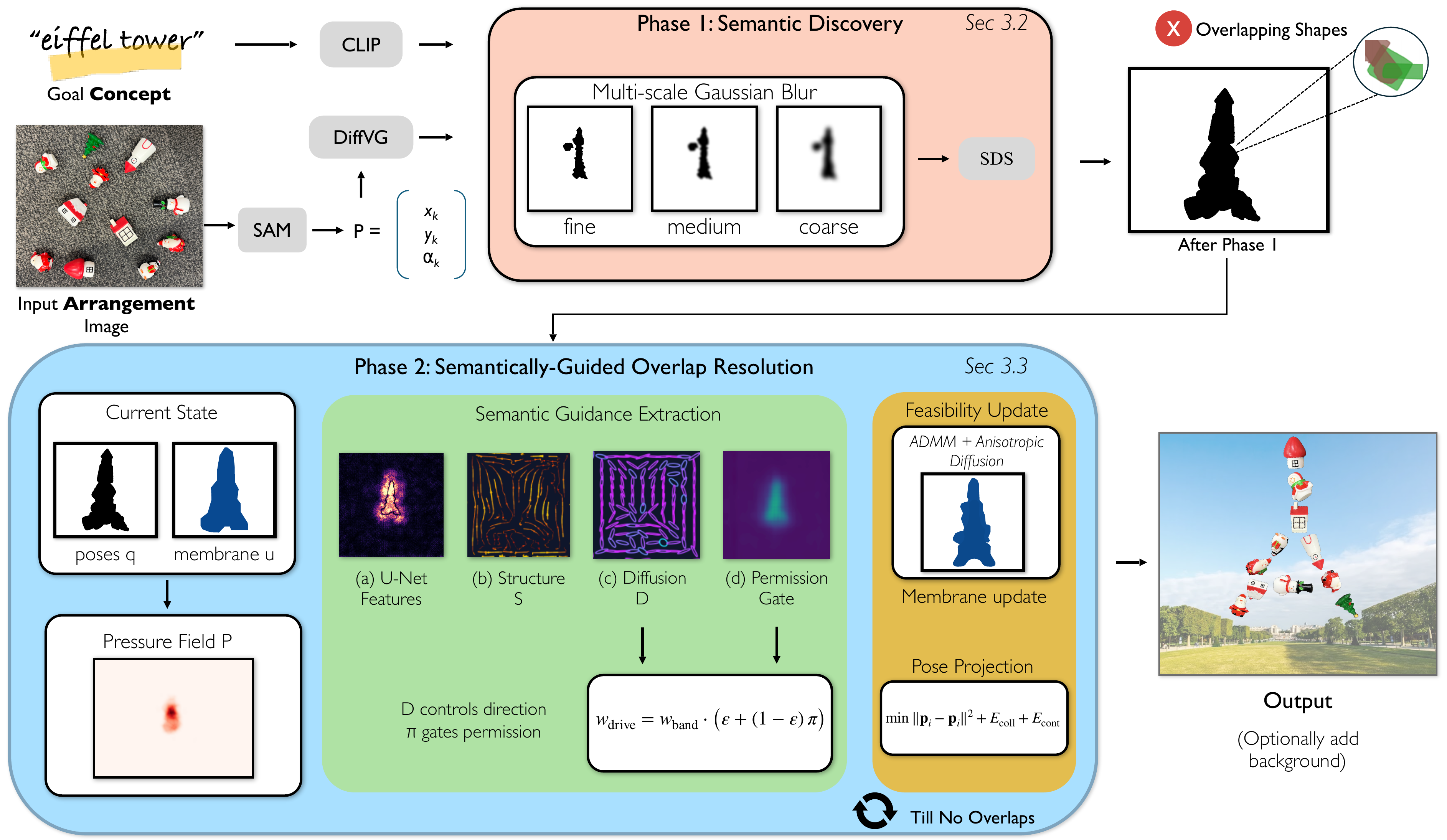}
    \caption{\textbf{ShapeShift pipeline.} Given input shapes and a text prompt, \textbf{Phase~1} (Sec.~\ref{sec:phase1}) uses Score Distillation Sampling with multi-scale Gaussian blur to discover semantically coherent arrangements, tolerating overlaps during optimization. \textbf{Phase~2} (Sec.~\ref{sec:phase2}) restores feasibility while preserving semantic structure by alternating membrane updates with pose projection. From the current poses $q$, membrane $u$, and pressure field $P$, we extract semantic guidance: (a)~UNet features from a mid-resolution decoder block, (b)~structure tensor $\mathbf{S}$ encoding local orientation, (c)~diffusion tensor $\mathbf{D}$ enabling anisotropic expansion along coherent directions, and (d)~permission field $\pi$ restricting expansion to feature-consistent regions. These combine into the gated drive $w_{\mathrm{drive}}$, which guides membrane updates via ADMM \cite{admm}, followed by pose projection. Optionally, a relevant background can also be generated by SD-XL~\cite{sdxl} for adding more context}
    \label{fig:pipeline}
\end{figure*}

\begin{figure}[t]
    \centering
    \includegraphics[width=\columnwidth]{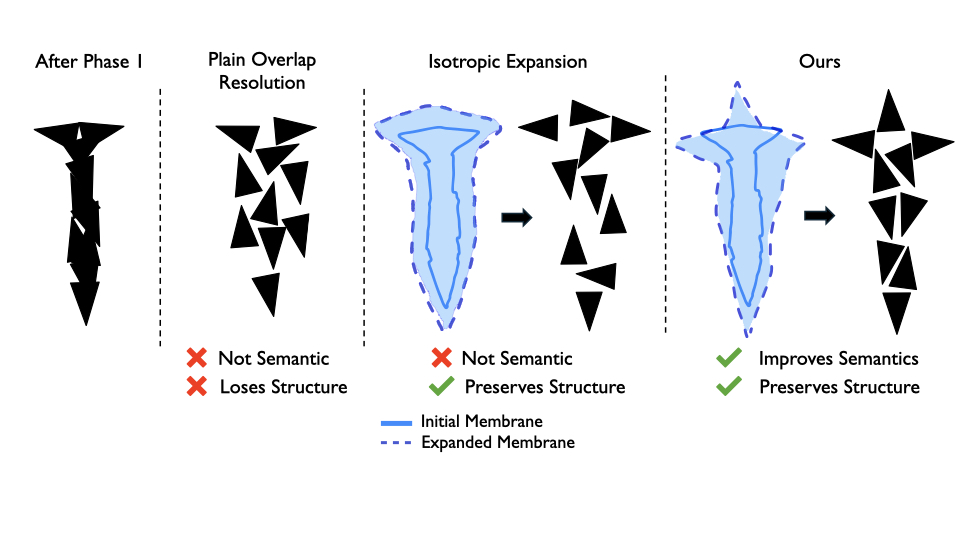}
    \caption{\textbf{Overlap resolution comparison.} Starting from an overlapping Phase~1 result (left), we compare three resolution strategies. \emph{Plain overlap resolution} separates shapes geometrically via minimum translation vectors, destroying the sword silhouette. \emph{Isotropic membrane expansion} provides containment but expands uniformly, weakening elongated structure. \emph{Ours} expands anisotropically guided by diffusion features, preserving both semantic fidelity and structural coherence.}
    \label{fig:baselines}
\end{figure}

\subsection{Problem Setup}
\label{sec:problem}

Given $N$ rigid 2D shapes $\{S_i\}_{i=1}^N$ and a text prompt $y$, we seek poses $q_i = (\mathbf{p}_i, \theta_i)$ comprising translation $\mathbf{p}_i \in \mathbb{R}^2$ and rotation $\theta_i \in [0, 2\pi)$ for each shape. We render arrangements differentiably via DiffVG~\cite{Li20} to obtain image $g(q)$. For geometric computations, we maintain per-shape occupancy fields $s_i: \Omega \to [0,1]$ on a grid $\Omega$ of resolution $128 \times 128$, where $s_i(\mathbf{x})$ indicates shape $i$'s soft coverage via signed-distance antialiasing.

Our goal is to find poses such that (i)~the rendered arrangement is semantically consistent with text prompt $y$, and (ii)~shapes do not overlap and remain within a bounded region.

\subsection{Phase 1: Semantic Discovery via SDS}
\label{sec:phase1}

Phase~1 uses Score Distillation Sampling (SDS)~\cite{poole2022dreamfusiontextto3dusing2d} to optimize poses toward a semantically meaningful arrangement. Given rendered image $x = g(q)$, SDS provides a gradient:
\begin{equation}
\nabla_q \mathcal{L}_{\mathrm{SDS}}
= \mathbb{E}_{t,\epsilon}\Bigl[
    w(t)\bigl(\hat{\epsilon}_\phi(z_t; y, t) - \epsilon\bigr)
    \frac{\partial x}{\partial q}
\Bigr],
\label{eq:sds}
\end{equation}
where $z_t$ is a noised latent encoding of $x$, $\hat{\epsilon}_\phi$ is the predicted noise conditioned on prompt $y$, and $w(t) = \sqrt{\alpha_t}\sigma_t$ weights by timestep.

\paragraph{Multi-scale blur augmentation.}
We apply Gaussian blur at multiple kernel sizes ($3\times3$, $5\times5$, $7\times7$) and average SDS gradients across scales, encouraging the model to respect both global structure and local detail.

\paragraph{Tolerating overlaps.}
We do \emph{not} enforce overlap constraints during Phase~1. Doing so traps optimization in scattered configurations; tolerating overlaps allows shapes to organize into semantically coherent layouts, left). Crucially, Phase~1 discovers arrangements \emph{expressible with the available primitives} (Fig.~\ref{fig:phase1_ablation})---our shapes form an incomplete basis that cannot realize arbitrary outlines---which Phase~2 then makes feasible.

\subsection{Phase 2: Semantically-Guided Feasibility Restoration}
\label{sec:phase2}

Phase~2 restores physical feasibility while preserving semantic structure. Standard geometric separation (e.g., minimum translation vectors) pushes shapes apart along locally shortest directions, which can be semantically destructive (Fig.~\ref{fig:baselines}). Our approach instead derives both directional bias and spatial gating from diffusion model features, enabling anisotropic membrane expansion along concept-coherent directions.

Phase~2 alternates: (i)~updating a phase-field membrane defining the feasible region, and (ii)~projecting poses toward non-overlap within the membrane. We terminate when overlap falls below $\tau_{\mathrm{stop}} = 0.5\%$ or after 500 iterations.

\subsubsection{Phase-Field Membrane}
\label{sec:membrane}

We represent the feasible region as a phase field $u: \Omega \to [0,1]$, where $u(\mathbf{x}) \approx 1$ indicates interior (feasible) and $u(\mathbf{x}) \approx 0$ indicates exterior. The level set $\{u = 0.5\}$ defines a soft boundary. We initialize $u$ as a tight envelope around the shape union (morphological dilation with radius 5 pixels) and evolve it to create space only where overlap pressure and semantic structure jointly permit.

\subsubsection{Pressure Field}
\label{sec:pressure}

We compute a pressure field driving membrane expansion:
\begin{equation}
P(\mathbf{x}) = \bigl[\max(0, \textstyle\sum_i s_i(\mathbf{x}) - 1)\bigr]^2
+ \bigl[\max(0, \textstyle\sum_i s_i(\mathbf{x}) - u(\mathbf{x}))\bigr]^2,
\label{eq:pressure}
\end{equation}
where the first term captures \emph{overlap pressure} and the second captures \emph{containment pressure} (shapes beyond the membrane). For early stopping, we track the discrete overlap percentage: the fraction of occupied pixels ($s > 0.1$) that are multiply occupied.

\subsubsection{Semantic Guidance from UNet Features}
\label{sec:tensor}

To determine \emph{how} the membrane should expand, we extract directional structure from diffusion model features. We compute intermediate features $\mathbf{f} \in \mathbb{R}^{C \times H \times W}$ from the UNet's second decoder block (\texttt{up\_blocks.1}, $64 \times 64$ resolution) during a forward pass on the current arrangement. Prior work shows such features encode spatially coherent semantic structure~\cite{Tang2023DIFT}.

From spatial gradients of channel-standardized features $\tilde{f}_c = (f_c - \mu_c)/\sigma_c$, we build a structure tensor:
\begin{equation}
\mathbf{S}(\mathbf{x}) = K_\sigma * \Bigl(\sum_{c=1}^{C} \nabla \tilde{f}_c \, \nabla \tilde{f}_c^\top\Bigr),
\label{eq:structure_tensor}
\end{equation}
where $K_\sigma$ is a Gaussian kernel ($\sigma = 0.5$ pixels). Eigendecomposition yields principal directions $\mathbf{e}_1, \mathbf{e}_2$ with eigenvalues $\lambda_1 \geq \lambda_2$, and coherence $c = (\lambda_1 - \lambda_2)/(\lambda_1 + \lambda_2 + \epsilon)$ measuring directional dominance.

We convert this to a diffusion tensor $\mathbf{D}$ controlling pressure propagation. The key insight: we want high diffusivity \emph{along} edges ($\mathbf{e}_2$ direction) and low diffusivity \emph{across} them ($\mathbf{e}_1$ direction):
\begin{equation}
\mathbf{D}(\mathbf{x}) = d_1 \,\mathbf{e}_1 \mathbf{e}_1^\top + d_2\, \mathbf{e}_2 \mathbf{e}_2^\top,
\label{eq:diffusion_tensor}
\end{equation}
where $d_2 = 1$ and $d_1 = 1/(1 + \beta c)$ with $\beta = 15$, yielding up to 16:1 anisotropy in coherent regions.

\subsubsection{Anisotropic Pressure Transport}
\label{sec:transport}

Given pressure $P$ and diffusion tensor $\mathbf{D}$, we compute transport potential $\varphi$ via anisotropic screened Poisson:
\begin{equation}
\alpha \varphi - \nabla \cdot \bigl(\mathbf{D}(\mathbf{x}) \nabla \varphi\bigr) = P(\mathbf{x}),
\label{eq:transport}
\end{equation}
with $\alpha = 0.01$, solved by conjugate gradient. The anisotropic diffusion causes pressure to escape preferentially along semantically coherent directions. At the membrane interface ($|u - 0.5| < 0.1$), we compute outward flux $w_{\mathrm{iface}} = -\nabla \varphi \cdot \mathbf{n}$ where $\mathbf{n} = \nabla u / \|\nabla u\|$, then smooth to obtain $w_{\mathrm{band}}$.

\subsubsection{Semantic Permission Field}
\label{sec:permission}

Transport determines where pressure \emph{wants} to escape; permission determines where expansion \emph{should} occur. We compute $\pi: \Omega \to [0,1]$ from feature consistency with the shape interior.

We maintain feature prototypes $\{\mathbf{f}_k^*\}$ sampled from high-occupancy regions and updated via EMA. For each location, we compute a soft nearest-neighbor score:
\begin{equation}
\mathrm{score}(\mathbf{x}) = \tau \log \sum_{k} \exp\bigl(\cos(\mathbf{f}(\mathbf{x}), \mathbf{f}_k^*) / \tau\bigr),
\label{eq:score}
\end{equation}
with $\tau = 0.1$. Permission is $\pi(\mathbf{x}) = \sigma((\mathrm{score} - \mathrm{score}_{\mathrm{med}})/T)$ where $\mathrm{score}_{\mathrm{med}}$ is the median over shape interiors (automatic calibration) and $T = 0.2$.

\paragraph{Gated expansion drive.}
The final drive combines transport and permission:
\begin{equation}
w_{\mathrm{drive}} = w_{\mathrm{band}} \cdot \bigl(\varepsilon + (1 - \varepsilon)\, \pi\bigr),
\label{eq:gated_drive}
\end{equation}
with floor $\varepsilon = 0.05$ preventing deadlock.

\subsubsection{Membrane Update via Alternating Direction Method of Multipliers (ADMM)}
\label{sec:admm}

We evolve $u$ using ADMM~\cite{admm} to enforce: (i)~containment $u \geq s$ where $s = \max_i s_i$; (ii)~bounds $u \in [0,1]$; and (iii)~area limits $A_{\min} \leq \int u \leq A_{\max}$ where $A_{\min}$ is total shape area plus 10\% margin and $A_{\max} = 2 A_{\min}$.

With auxiliary $z$ and scaled dual $y$, we iterate:
\begin{align}
\textbf{u-step:}&\quad
\bigl(\rho \mathbf{I} - \nabla\!\cdot\!(\mathbf{D}\nabla)\bigr) u^{k+1}
= \rho(z^k - y^k) + w_{\mathrm{drive}}, 
\label{eq:u_step}\\
\textbf{z-step:}&\quad
z^{k+1} = \Pi_{\mathcal{C}}\bigl(u^{k+1} + y^k\bigr),
\label{eq:z_step}\\
\textbf{y-step:}&\quad
y^{k+1} = y^k + (u^{k+1} - z^{k+1}),
\label{eq:y_step}
\end{align}
where $\rho = 1.0$ and $\Pi_{\mathcal{C}}$ projects onto the constraint set.



\subsubsection{Pose Projection}
\label{sec:projection}

Given the updated membrane, we project poses toward feasibility by solving:
\begin{equation}
\min_{\{q_i\}}
\sum_i \left[
\frac{\|\mathbf{p}_i - \mathbf{p}_i^{(t)}\|^2}{2\tau_p}
+
\frac{r_i^2 \, d_\theta(\theta_i,\theta_i^{(t)})^2}{2\tau_\theta}
\right]
+ w_{\mathrm{coll}} E_{\mathrm{coll}}
+ w_{\mathrm{cont}} E_{\mathrm{cont}},
\label{eq:projection}
\end{equation}
where $r_i$ is shape $i$'s circumradius, $d_\theta$ is wrapped angular distance, and we use $\tau_p = \tau_\theta = 0.8$, $w_{\mathrm{coll}} = 10$, $w_{\mathrm{cont}} = 5$.

\paragraph{Collision energy.}
We use differentiable Minkowski-sum penetration~\cite{Minarcik2024MinkowskiPenalties}. For convex shapes, penetration depth along direction $\mathbf{d}$ is $g_{ij}(\mathbf{d}) = h_{S_i}(\mathbf{d}) + h_{S_j}(-\mathbf{d}) - \mathbf{d} \cdot (\mathbf{p}_j - \mathbf{p}_i)$, where $h_S$ is the support function approximated via log-sum-exp over vertices. We minimize over SAT-complete directions and apply convex decomposition for non-convex shapes. The collision energy is $E_{\mathrm{coll}} = \sum_{i<j} \mathrm{softplus}(g_{ij}/\sigma_g)^2$ with $\sigma_g = 2$. Containment energy $E_{\mathrm{cont}}$ penalizes boundary points where $u < s$.
\section{Results}
\label{sec:results}

We evaluate ShapeShift through quantitative metrics, baseline comparisons, and human studies. 
Our experiments address the following research question: 1) Should semantic guidance and feasibility constraint resolution be coupled?  





\subsection{Ablation Study}
\label{sec:ablation}

\subsubsection{Phase~1 Ablation}
To evaluate the benefit of constraint-aware semantic guidance in Phase~1, 
we compare the performance with the following two simple initial shape generation methods:

\begin{itemize}[leftmargin=*, itemsep=0pt]
\item \textbf{Prompt silhouette container:} Generate a target silhouette via Stable Diffusion, binarize, and use as the initial membrane for Phase~2 projection.

\item \textbf{Generic circular container:} Initialize the membrane as a circle and run Phase II. 
\end{itemize}

\begin{figure*}[t]
    \centering
    \includegraphics[width=0.9\textwidth]{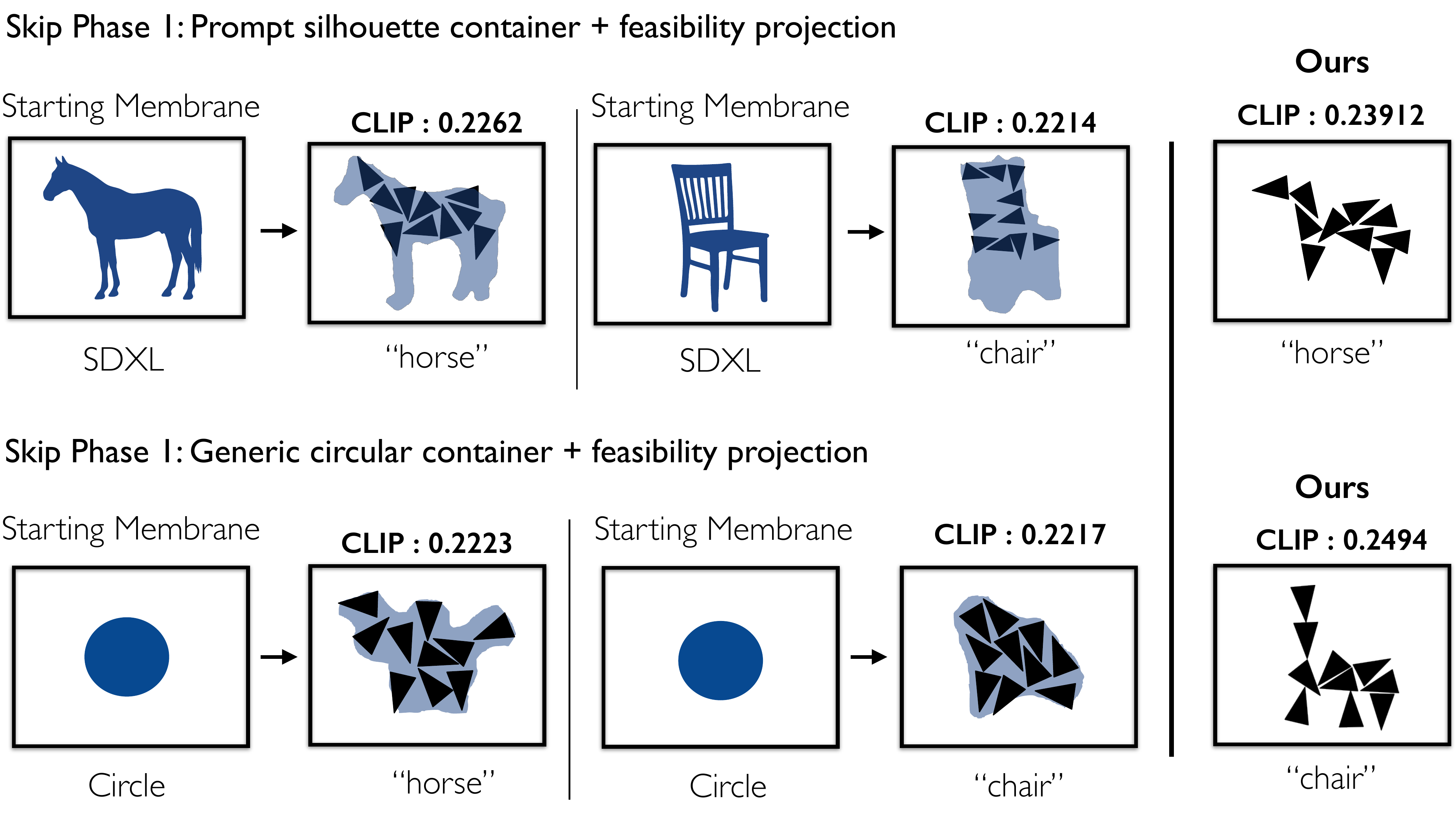}
    \caption{\textbf{Why Phase~1 is necessary.} We compare approaches that skip Phase~1 semantic discovery against our full pipeline (rightmost column). \emph{Top:} Using a prompt-generated silhouette (SDXL) as target container yields CLIP scores of 0.226 and 0.221---shapes cannot realize arbitrary outlines like thin horse legs. \emph{Bottom:} Generic circular containers produce efficient packing but no semantic structure (CLIP: 0.222, 0.217). Our full pipeline achieves higher scores (0.239, 0.249) because Phase~1 discovers arrangements \emph{expressible with available pieces}, which Phase~2 then makes feasible.}
    \label{fig:phase1_ablation}
\end{figure*}

As shown in Fig~\ref{fig:phase1_ablation}, both produce substantially lower CLIP scores (0.217--0.226) compared to our full pipeline (0.239--0.249). The silhouette approach fails because shapes form an incomplete basis---they cannot realize arbitrary outlines (thin horse legs from blocky pieces). The generic container fails because overlap resolution alone has no semantic direction. 

This study validates that \textbf{an initial membrane guided by both semantic and feasibility constraints leads to semantically higher quality results.}



\subsubsection{Phase~2 Ablation}
To evaluate how useful semantically guided constraint resolution is during Phase~2, 
we compare our full method against two alternatives:
\begin{itemize}[leftmargin=*, itemsep=0pt]
    \item \textbf{Plain overlap resolution}: MTV-based separation without any membrane---shapes are pushed apart along minimum-distance directions.
    \item \textbf{Isotropic membrane}: A membrane that expands uniformly wherever pressure demands, without directional or spatial bias from diffusion features.
\end{itemize}

\begin{table}[t]
\centering
\caption{\textbf{Phase~2 Ablation}
We compare feasibility restoration strategies on 240 arrangements (8 trials $\times$ 30 concepts). Plain resolution destroys semantic structure despite achieving the lowest overlap. Semantic guidance preserves structure at the cost of slightly higher residual overlap. Statistical significance: paired $t$-test, $p < 0.01$.}
\label{tab:ablation}
\resizebox{\columnwidth}{!}{%
\begin{tabular}{lcc}
\toprule
\textbf{Method} & \textbf{Overlap (\%)}$\downarrow$ & \textbf{CLIP Score}$\uparrow$ \\
\midrule
Phase 1 only (no resolution) & $32.4 \pm 2.9$ & $0.251 \pm 0.013$ \\
\midrule
Plain Overlap Resolution & $\mathbf{0.2 \pm 0.3}$ & $0.231 \pm 0.021$ \\
Isotropic Membrane & $0.8 \pm 0.3$ & $0.234 \pm 0.015$ \\
\textbf{Semantic Guidance (Ours)} & $0.9 \pm 0.4$ & $\mathbf{0.244 \pm 0.018}$ \\
\bottomrule
\end{tabular}%
}
\end{table}

The results in Table~\ref{tab:ablation} shows that
the plain MTV-based resolution achieves the lowest overlap (0.2\%) but produces the lowest CLIP score (0.231)---an 8\% degradation from Phase~1. Shapes scatter in geometrically optimal but semantically destructive directions. Isotropic expansion improves slightly (0.234 CLIP) by providing containment, but uniform expansion still weakens directional structures.

Semantic guidance achieves the highest CLIP score (0.244)---recovering most of the Phase~1 semantic quality while achieving feasibility. The 5.6\% improvement over plain resolution is statistically significant ($p < 0.01$).

\begin{figure*}[t]
    \centering
    \includegraphics[width=\textwidth]{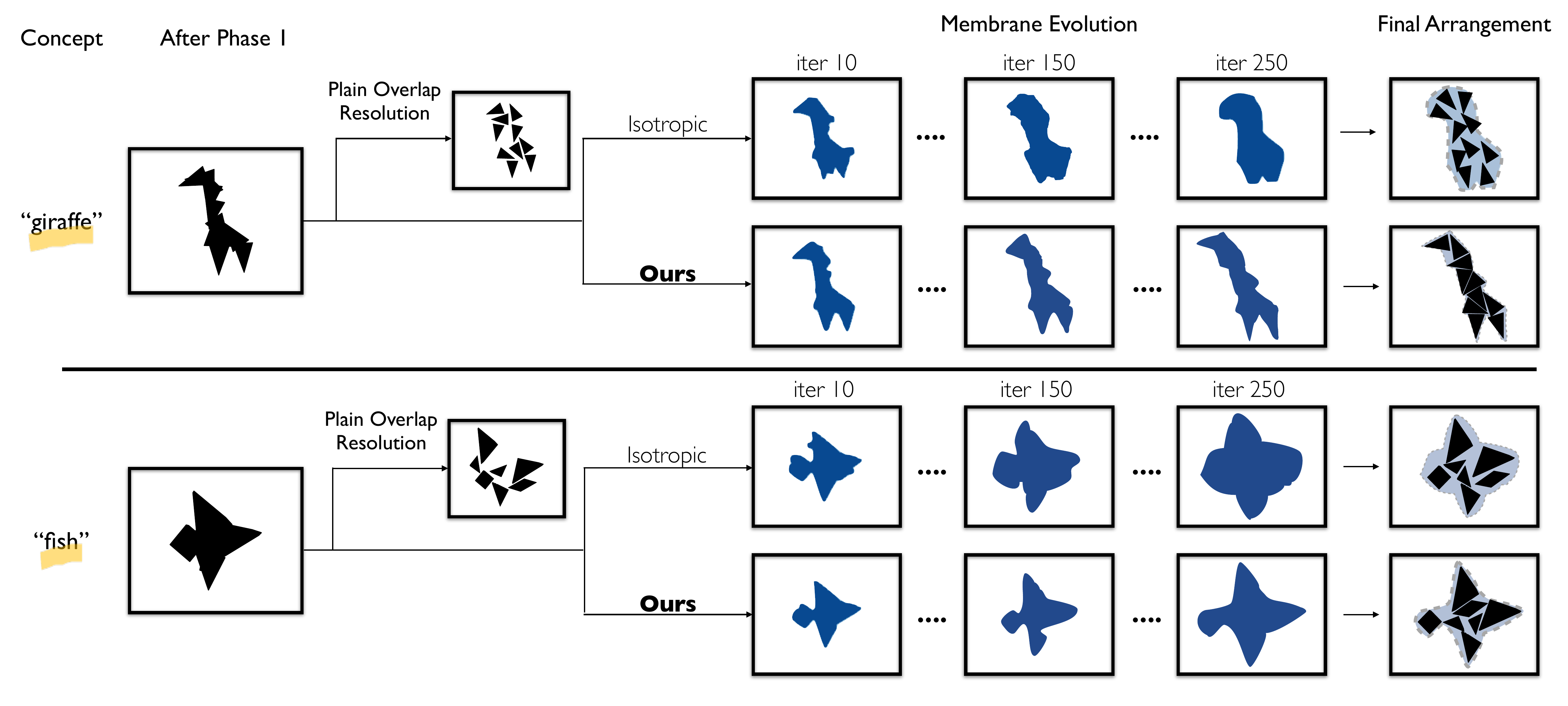}
    \caption{\textbf{Membrane evolution: isotropic vs.\ semantic guidance.} For two concepts (giraffe, fish), we show Phase~1 output, plain overlap resolution, and membrane evolution at iterations 10, 150, 250. Isotropic expansion bulges uniformly, distorting elongated silhouettes. Semantic guidance expands along body axes, preserving structure. Blue regions: membrane interior $\{u > 0.5\}$.}
    \label{fig:membrane_evolution}
\end{figure*}

Fig.~\ref{fig:membrane_evolution} visualizes the differences. For elongated concepts (giraffe, fish), isotropic expansion bulges uniformly, distorting silhouettes. Semantic guidance expands preferentially along body axes, preserving elongated structure.

These results validate that \textbf{constraint resolution with semantic guidance produces significantly more recognizable outputs.}


\subsection{Comparison with Generative Approaches}
\label{sec:baselines}

\begin{figure*}[t]
    \centering
    \includegraphics[width=\textwidth]{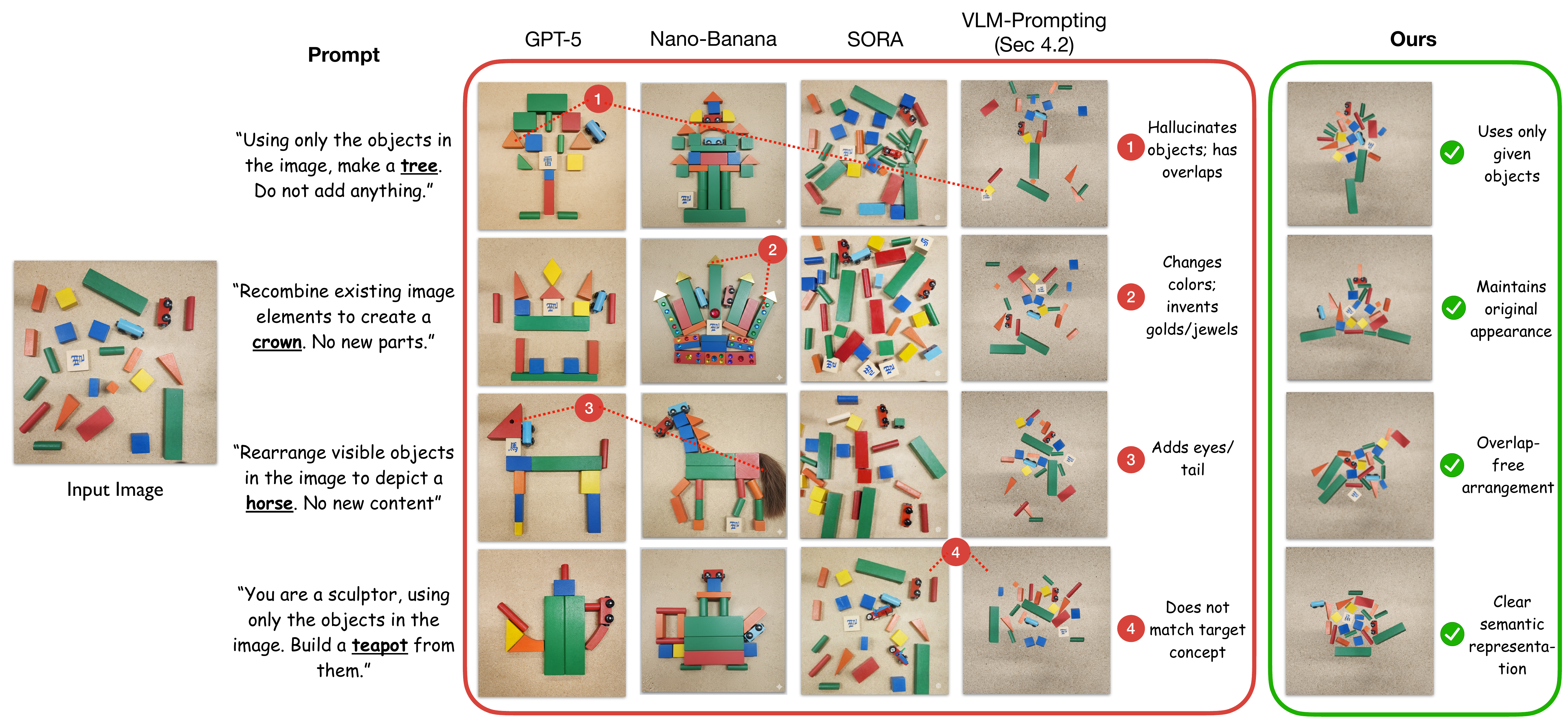}
    \caption{\textbf{Comparison with generative models and VLM planning.} Given input objects (leftmost column) and text prompts (tree, crown, horse, teapot), we compare arrangements from text-to-image models (GPT-4o, Nano Banana Pro), video generation (Sora), VLM coordinate planning, and our method (rightmost). Generative models hallucinate objects, modify colors, and produce overlaps. VLM planning achieves coarse layouts but struggles with precise positioning. Only ShapeShift maintains object identity while producing overlap-free, semantically coherent arrangements.}
    \label{fig:generative_comparison}
\end{figure*}

We qualitatively compare the proposed approach against three state-of-the-art generative approaches: text-to-image generation (GPT-4o~\cite{openai2024gpt4technicalreport}, Nano Banana Pro~\cite{geminiteam2025geminifamilyhighlycapable}), video generation (Sora~\cite{liu2024sorareviewbackgroundtechnology}). We also included a naive VLM-based spatial planning designed as follows: We prompted GPT-4V with: (1)~the input image with a $10 \times 10$ coordinate grid overlay, (2)~labeled object identifiers (A--J), (3)~the target concept, and (4)~instructions to output $(x, y, \theta)$ for each object while avoiding overlaps. When rendered results contained overlaps, we provided visual feedback and requested refinement (up to 5 iterations).

Fig.~\ref{fig:generative_comparison} presents results across four target concepts. Despite explicit instructions to use only the given objects without modification, all generative baselines exhibit systematic failures such as object hallucination and modification as well as violations of physical validity. 


These failures arise from a fundamental mismatch: pixel-based generation optimizes visual plausibility, not physical validity.  ShapeShift addresses this through explicit geometric control---each object maintains identity and boundaries---combined with semantic guidance from diffusion features.
\subsection{Human Evaluation}

CLIPScore can provide a general understanding of semantic alignment, but it can be overfit to and noisy. To truly evaluate semantic alignment, we conducted a human evaluation study using Qualtrics surveys on Prolific. We recruited 30 participants who each viewed images generated by various methods and tried to identify the concept the arrangements depicted. There were four choices, so a random guess would result in 25\% accuracy. We report our results in Table~\ref{tab:userstudy}.  While these results follow the general trend of CLIPScores in Table~\ref{tab:ablation}, the difference between our approach and baselines in human identification accuracy and CLIPScore is massively different. Our approach resulted in over a 30\% improvement in identification accuracy over Isotropic Expansion whereas this was only about a 4\% improvement in CLIPScore. This highlights the differences in automatic metrics versus true human judgment.

\begin{table}[h]
\centering 
\caption{\textbf{Human-Rater Results.} We report the accuracy with which human raters were able to identify the concept in images generated by various methods.}
\resizebox{\columnwidth}{!}{%
\begin{tabular}{lc}
\toprule
Method & \textbf{Identification Accuracy}  \\
\midrule
Plain Overlap Resolution & $31.85\% \pm 2.32$ \\
Isotropic Expansion & $32.15\% \pm 2.35$ \\
\textbf{Semantic Guidance (Ours)} & $\mathbf{43.75\% \pm 2.48}$  \\
\bottomrule
\end{tabular}
}
\label{tab:userstudy}
\end{table}
\label{sec:human_eval}

\subsection{Qualitative Results}
\label{sec:qualitative}

\begin{figure*}[p]
    \centering
    \includegraphics[width=0.8\textwidth]{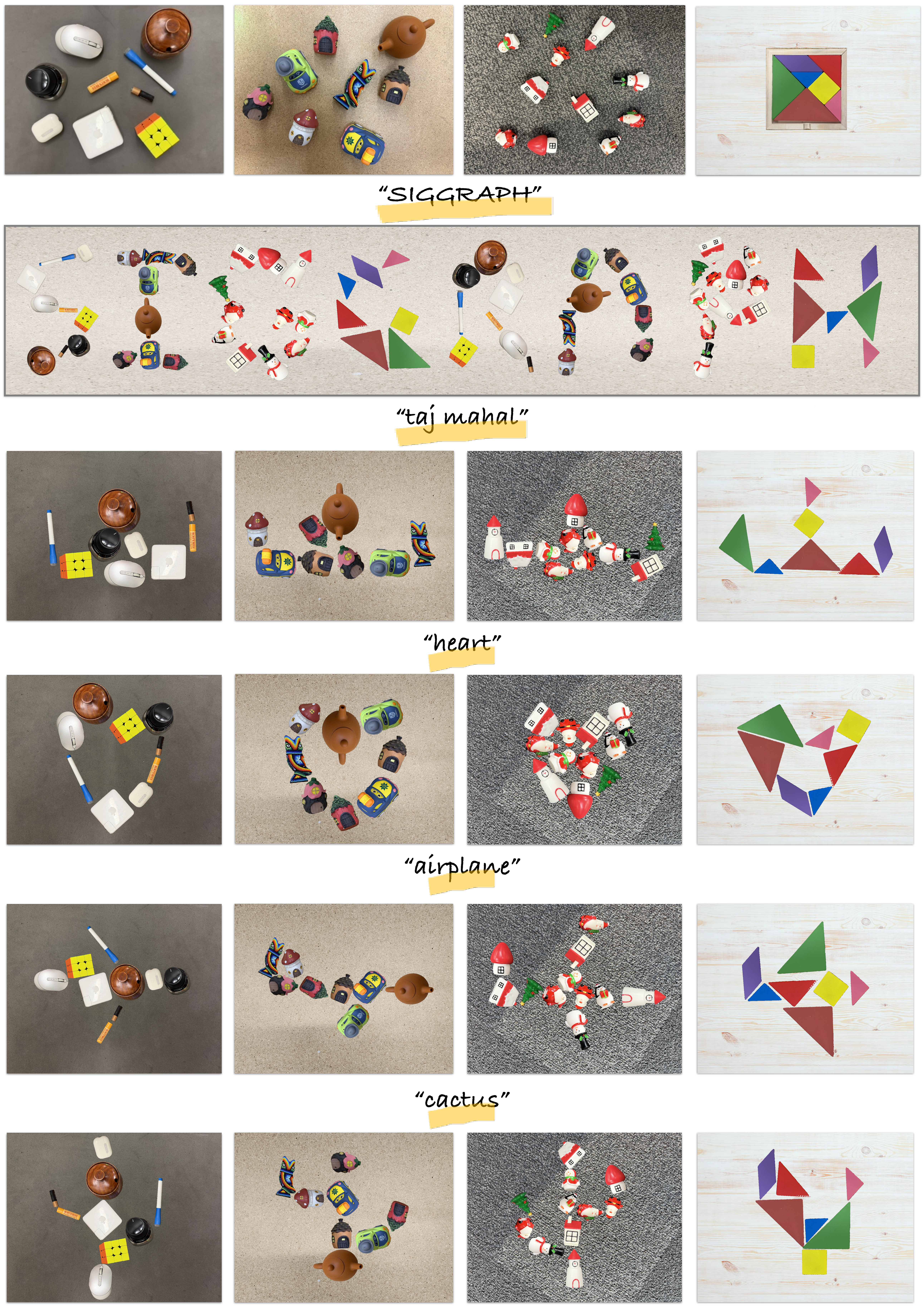}
    \caption{\textbf{Qualitative results across diverse object sets and concepts.} Each row shows a target concept; columns show different object sets (household items, toy vehicles, figurines, tangram pieces). ShapeShift produces semantically recognizable, overlap-free arrangements across all combinations, adapting to each object vocabulary. All arrangements use every piece exactly once with no overlaps. Best viewed zoomed.}
    \label{fig:qualitative_grid}
\end{figure*}

Fig.~\ref{fig:qualitative_grid} presents results across four diverse object sets and six target concepts. The object sets span geometric complexity: tangram pieces (7 convex polygons), toys (8 varied shapes), decorative figurines (11 complex silhouettes), and office tabletop items (9 distinct objects). Target concepts range from abstract symbols (heart) to recognizable objects (airplane, cactus) to complex structures (Taj Mahal, SIGGRAPH text).

Across all combinations, ShapeShift produces arrangements that are both semantically recognizable and physically valid, as opposed to other methods which struggle to adapt to complex, specific objects (Fig.~\ref{fig:generative_comparison}). ShapeShift adapts to different object vocabularies: tangram arrangements leverage geometric simplicity, while figurine arrangements exploit semantic affordances. This flexibility emerges from asset-conditioned discovery---Phase~1 finds arrangements expressible with available pieces rather than forcing pieces into predetermined templates.
\section{Discussion}
\label{sec:discussion}

\ssname demonstrates that semantic preservation and geometric validity need not be competing objectives. By guiding overlap resolution with structure extracted from diffusion model features, our approach eliminates overlaps while preserving the semantic arrangements discovered in Phase~1. The quantitative results (Table~\ref{tab:ablation}) and human evaluation (Table~\ref{tab:userstudy}) together validate this design: semantically-guided resolution substantially outperforms purely geometric alternatives in both automated metrics and human recognition.

\paragraph{Why geometric resolution fails.}
The performance gap between our method and geometric baselines (43.75\% vs.\ 31.85-32.15\% human accuracy; 0.2444 vs.\ 0.2312--0.2341 CLIP score) reveals a fundamental limitation of geometry-only approaches. Plain MTV-based resolution pushes shapes apart along minimum-distance directions, scattering arrangements in semantically arbitrary ways. Isotropic membrane expansion improves upon this by providing containment, but still treats all expansion directions as equivalent. In both cases, a sword may be widened rather than extended; a tower may be shortened rather than narrowed. These geometrically valid solutions destroy the semantic structure that makes arrangements recognizable. Our semantically-guided approach addresses this by biasing resolution toward directions and regions that preserve concept identity.

\paragraph{How semantic overlap resolution works.}
Our semantic guidance extracts structure from UNet features to inform membrane expansion. The diffusion tensor identifies local anisotropy---directions along which semantic content varies least---while feature consistency determines which regions are appropriate for expansion. These operate as a unified framework: the key empirical finding is that semantic guidance as a whole substantially outperforms both plain overlap resolution and isotropic expansion, enabling overlap resolution that preserves rather than destroys semantic structure.


\paragraph{Limitations.}
Several limitations warrant discussion. The framework is restricted to 2D arrangements with position and orientation parameters---extending to 3D would require volumetric representations and more complex overlap reasoning. Performance depends on the diffusion model's semantic understanding; concepts poorly represented in its training data yield weaker guidance. The system also struggles with underconstrained problems (too few objects to represent a concept) or prompts describing inherently dynamic scenes that static arrangements cannot capture.

\paragraph{Failure cases.}
We observe degraded performance in two scenarios. When objects are highly heterogeneous in scale, the pressure field may be dominated by large objects, causing small objects to be displaced disproportionately. When the target concept requires precise relative positioning (e.g., facial features), the continuous optimization may not converge to the exact configuration needed for recognition. Future work could address these through scale-aware pressure normalization and discrete refinement stages.

\paragraph{Toward robotic assembly.}
A natural extension of \ssname is generating target configurations for robotic pick-and-place tasks. Our output---poses $(x, y, \theta)$ for each object---directly specifies goal states for a manipulation pipeline. Prior work on language-guided rearrangement~\cite{liu2022structdiffusion, kapelyukh2024dream2realzeroshot3dobject} has demonstrated the challenge of bridging high-level semantic instructions to precise spatial goals. \ssname could serve as a semantic planner in such pipelines, translating natural language into concrete spatial targets that a motion planner can execute. The overlap-free guarantee is particularly relevant for physical execution, where intersecting goal poses would be infeasible. Extending this framework to 3D, incorporating grasp feasibility, and handling real-world sensing uncertainty are directions for future work.
\section{Conclusion}
\label{sec:conclusion}

We presented \textbf{\ssname}, a framework for synthesizing semantically meaningful, overlap-free arrangements of rigid objects from text. Our two-phase approach---semantic discovery via SDS, then semantically-guided feasibility restoration---addresses the tension between geometric validity and semantic preservation.
%
The key empirical finding is that geometric resolution without semantic guidance destroys the meaning of the structure, dropping human recognition from 43.8\% to 31.9\%.
By extracting directional and spatial priors from UNet features, we preserve semantic structure while achieving physical feasibility.


\bibliographystyle{ACM-Reference-Format}
\bibliography{main}

\end{document}